\documentclass[runningheads]{llncs}
\usepackage{graphicx}
\usepackage{amsmath,amssymb,amsfonts} %

\usepackage[utf8]{inputenc} %
\usepackage[T1]{fontenc}    %

\usepackage[font=small,skip=4pt]{subcaption}
\usepackage[font=small,skip=4pt,tableposition=top]{caption}

\usepackage[numbers,sort,compress]{natbib} 

\usepackage{booktabs}  
\usepackage{multirow}
\usepackage{microtype} 
\usepackage{nicefrac}  
\usepackage{newtxmath} 

\usepackage{hyperref}

\usepackage[dvipsnames]{xcolor}         
\usepackage{graphicx}
\usepackage{pifont}%

\usepackage{comment}
\usepackage[width=122mm,left=12mm,paperwidth=146mm,height=193mm,top=12mm,paperheight=217mm]{geometry}

\hypersetup{
     colorlinks   = true,
     citecolor    = blue
}

\begin{document}
\mainmatter

\title{End-to-end Tracking with a Multi-query Transformer}
\titlerunning{\em MQT}
\authorrunning{Korbar and Zisserman}

\author{
    Bruno Korbar*\and 
    Andrew Zisserman 
}

\institute{Visual Geometry Group\\ University of Oxford \\
*\email{korbar@robots.ox.ac.uk}
}

\maketitle

\begin{abstract}
Multiple-object tracking (MOT) is a challenging task that requires simultaneous reasoning about location, appearance, and identity of the objects in the scene over time. Our aim in this paper is to move beyond tracking-by-detection approaches, that perform well on datasets where the object classes are known, to class-agnostic tracking that performs well also for unknown object classes.
To this end, we make the following three contributions: first, we introduce {\em semantic detector queries} that enable an object to be localized by specifying its approximate position, or its appearance, or both; second, we use these queries within an auto-regressive framework for tracking, and propose a multi-query tracking transformer (\textit{MQT}) model for simultaneous tracking and appearance-based re-identification (reID) based on the transformer architecture with deformable attention. This formulation allows the tracker to operate in a class-agnostic manner, and the model can be trained end-to-end; finally, we demonstrate that \textit{MQT} performs competitively on standard MOT benchmarks,  outperforms all baselines on generalised-MOT, and generalises well to a much harder tracking problems such as tracking any object on the TAO dataset.
\end{abstract}

\section{Introduction}

The objective of this paper is \textit{multi-object tracking} (MOT) --
the task of determining the spatial location of multiple objects over
time in a video. This is a very well researched area and, broadly, two
approaches are dominant: the first is {\em tracking-by-detection},
where a strong {\em object category detector} is trained for the object
class of interest, for example a person or a car. This approach proceeds in two steps: the detector is first
applied independently on each frame, and in the second step, the tracking task 
reduces to the data association of grouping these detections over time (over the frames in this
case). Examples of this approach include~\cite{bergmann2019bells, zhou2020tracking, wang2020towards, chen2018realtime}. The second approach
is {\em class agnostic tracking} where {\em any} object can be tracked. The object of interest is
specified by a bounding box or segmentation in one frame, and the task
is then to track that object through the other frames. Examples of
this approach include~\cite{bertinetto2016fully, danelljan2017eco, held2016learning}.  

\begin{center}
    \centering
    \includegraphics[width=\linewidth]{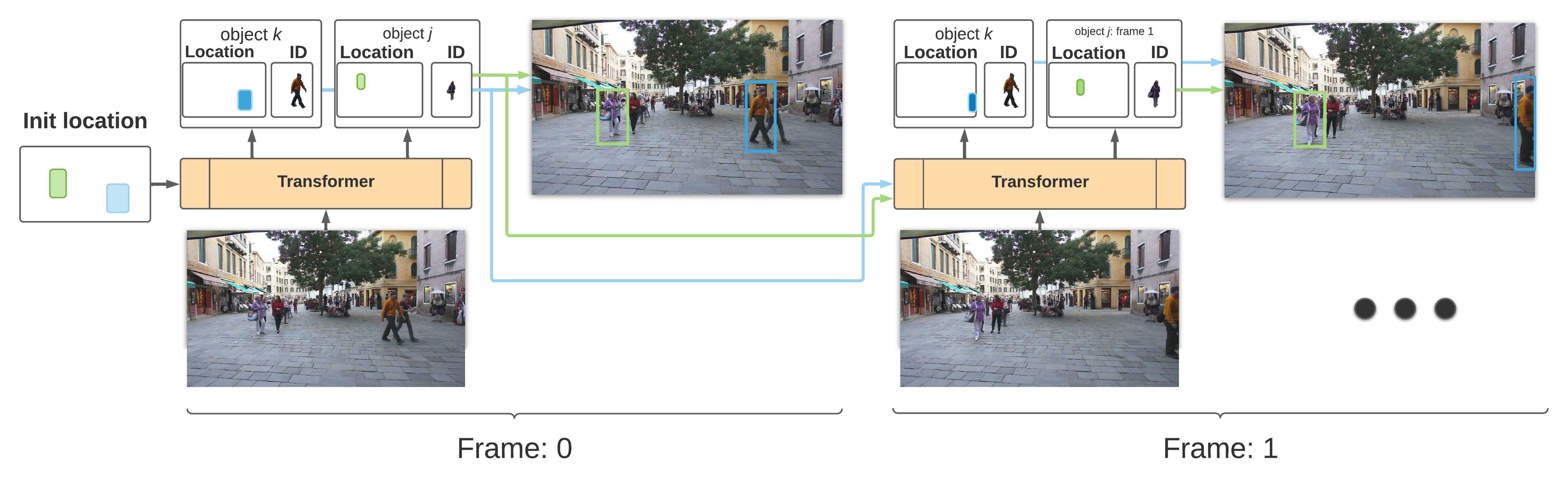}
    \captionsetup{type=figure}
    \captionof{figure}{\small An overview of the functionality of the multi-query tracking transformer (\textit{MQT}). Each frame generates location and appearance embeddings of the target object. These embeddings are used as queries for the subsequent frame. By propagating information between frames in this simple manner the object is tracked over time through the video.}
    \label{fig:splash}

\end{center}%

\noindent The tracking-by-detection approach generally outperforms class-agnostic models at the moment, 
but the approaches often suffer from overly complex processing pipelines 
(using multiple separately trained models for each step) and they rely
on prior knowledge of the object class of interest. More importantly the 
detection model and the data-association model are in tension: with
one model trained to tolerate 
object class variations (to better detect all instances of the same class), whilst the
other is trained to maximise discrimination of two instances of the
same class (to prevent identity switching). Such models are generally not trained end-to-end. Lastly, such models are highly specific -- the results of these trackers often don't generalise well to the more general tracking scenario~\cite{achal2020TAO}.

\noindent In this paper, we present a {\em class-agnostic} tracker that can be trained end-to-end, but also build on the
lessons of a strong object category detector. To this end we base the tracker
on the DETR object category detector~\cite{carion2020DETR},
using a transformer-detector modified in such a way
that it can attend to multiple objects locations and identities
simultaneously. We introduce dual `object-specific location' and `identity'
encodings (dubbed \textit{semantic queries}) which allow the model to
selectively focus on the {\em location} or {\em appearance} of objects we want to
track, irrespective of their classes. These object-specific embeddings
enable the model to be optimized {\em jointly} for track prediction and
re-identification by training in a class agnostic manner. 
In this way we achieve a single model class-agnostic tracker that performs competitively on several MOT benchmarks~\cite{MOT16and17, MOT20}, and can outperform all previous work on the class agnostic-MOT task~\cite{bai2021gmot40} where the class-prior is not known. Lastly, we show that the tracker trained in this way can also generalise well to tracking task such as TAO~\cite{achal2020TAO}, where the categories and number of tracking targets is far more general than on MOT benchmarks.



To summarise, we make the following three contributions:
First, we introduce the concept of semantic detector-queries and show their effectiveness for multi-object tracking.
Second,  we design a transformer-based class-agnostic tracking model around semantic detector-queries that is capable of simultaneous detection and re-identification of multiple objects in the scene.
Finally, we achieve competitive results on various MOT benchmarks~\cite{MOT16and17, MOT20} where object identity is used, demonstrate state-of-the-art class-agnostic performance on generalized MOT~\cite{bai2021gmot40}, and show the potential of the model to generalise to even harder tracking tasks on TAO~\cite{achal2020TAO}.

\section{Related work}
\label{sec:related work}
To put this work into context, we compare it to the modern tracking approaches that use a similar tracking paradigm to ours. There are, of course, many other tracking approaches (e.g.\ tracking-by-segmentation~\cite{osep2018track, voigtlander2019mots, zu2020segment, gedas2020maskprop}) that are not as closely related to our method. 

\noindent\textbf{Tracking by detection} approaches form trajectories by associating detections over time~\cite{zhou2020tracking, wang2020towards, chen2018realtime}. A common way of representing the data-association problem is to view it as a graph, where each detection is a node linked by possible edges and formulating it as a maximum-flow problem~\cite{berclaz2011multiple} with distance based~\cite{pirsiavash2011globally, zhang2008global} or learned costs~\cite{lealtaix2014learning}. Alternative formulations use association graphs~\cite{ma2018customized}, learned models based on motion models~\cite{kim2015multiple}, or a completely learned graph-neural-network~\cite{braso2020learning}. A common issue with graph-based approaches is the high optimization cost that doesn't necessarily translate to better performance. 

\noindent Detections can also be associated by modelling \textit{motion} directly~\cite{alahi2016social, leal2011everybody}. Pre-deep learning approaches often rely on assumptions of constant motion~\cite{choi2010Multiple, andriyenko2011multitarget} or existing models of human behaviour~\cite{scovanner2009learning, pellegrini2009youll, yamaguchi2011who}, whilst more modern approaches attempt to learn the motion models directly from the data~\cite{lealtaix2014learning}. Our model doesn't model motion explicitly, although, we do rely on the assumption of small motion within frames to account for appearance similarity.

\noindent  \textbf{Tracking by appearance} methods use increasingly powerful image-representations to track objects based on the similarities produced by either Siamese-networks~\cite{lealtaix2014learning, shuai2021siammot}, learned reID features~\cite{ristani2018features}, or other alternative methods~\cite{chen2018realtime, chu2019famnet, pang2021qdtrack}.

\noindent \textbf{Tracking by regression} \textit{refines} (instead of detecting) the bounding box of the current frame by regressing the current bounding box given the bounding box at the previous frame~\cite{bergmann2019bells,braso2020learning, feichenhofter2017detect, zhou2020tracking}. As these models usually lack information about the object identity or relative track location, additional reID and motion models~\cite{bergmann2019bells, feichenhofter2017detect, zhou2020tracking} or graph methods~\cite{braso2020learning} are necessary to achieve competitive performance. Our model falls roughly in this category, although we show that it can learn reID information directly from data.

\noindent\textbf{Tracking with transformers} uses aspects
of the transformer architecture~\cite{Vaswani17Transformers},  such as self-attention and set-prediction~\cite{carion2020DETR, zhu2021ddetr}.
The \textit{Trackformer}, a transformer tracker proposed by~\citet{trackformer}, is the closest approach to ours, employing largely the same architecture model, but use class information for tracking and do not employ semantic queries. The TransTrack model~\cite{sun2021transtrack} operates in the same way as~\cite{trackformer} but with a different underlying backbone. MOTR~\cite{zeng2021motr} extends this framework by adding a ``query-interaction-module" to reason about track-queries over time. \citet{yu2021relationtrack} leverage the importance of semantically-decoupled embeddings. They employ the ``global context disentangling unit" to separate the final layer output of a backbone CNN directly to semantic embeddings; we on another hand, do it in the transformer decoder. TrackCenter model~\cite{xu2021transcenter} introduces two key improvements: pixel-level dense-queries, and semantically-decoupled representation learning via model separation. TransMOT~\cite{chu2021transmot} utilises transformers in a different way, by introducing a spatio-temporal graph transformers for post detection data-association. MeMOT~\cite{cai2022memot} introduces a memory module on top of the transformer encoder to further boost performance. Note that \textit{none} of these works can be generalised to GMOT or TAO tasks, as they are tracking-by-detection approaches and cannot
be used for class-agnostic tracking. For more in-depth comparison to most-similar works please refer to the supplementary material.

\noindent \textbf{Class-agnostic tracking} leverages powerful appearance embeddings to track objects based the similarity of the embeddings. The method does not leverage class information explicitly. These models often use a form of a Siamese architecture to learn a patch-based matching function~\cite{lealtaix2014learning, held2016learning, danelljan2017eco, tao2016siamese, bertinetto2016fully, leal2016learningsiamese, shuai2021siammot}. However, even if the model is in principle capable of class-agnostic inference, models such as~\cite{shuai2021siammot} are not fully class-agnostic, as they require class information for successful training of their tracker in the form of an object detection loss (that requires ground-truth class information for every object in the training triplet). Our work differs in that it does not require this
explicit object class labelling.

\section{Multi-query transformer for tracking} \label{sec:method}

\begin{figure}[t]
\centering
\includegraphics[width=\linewidth]{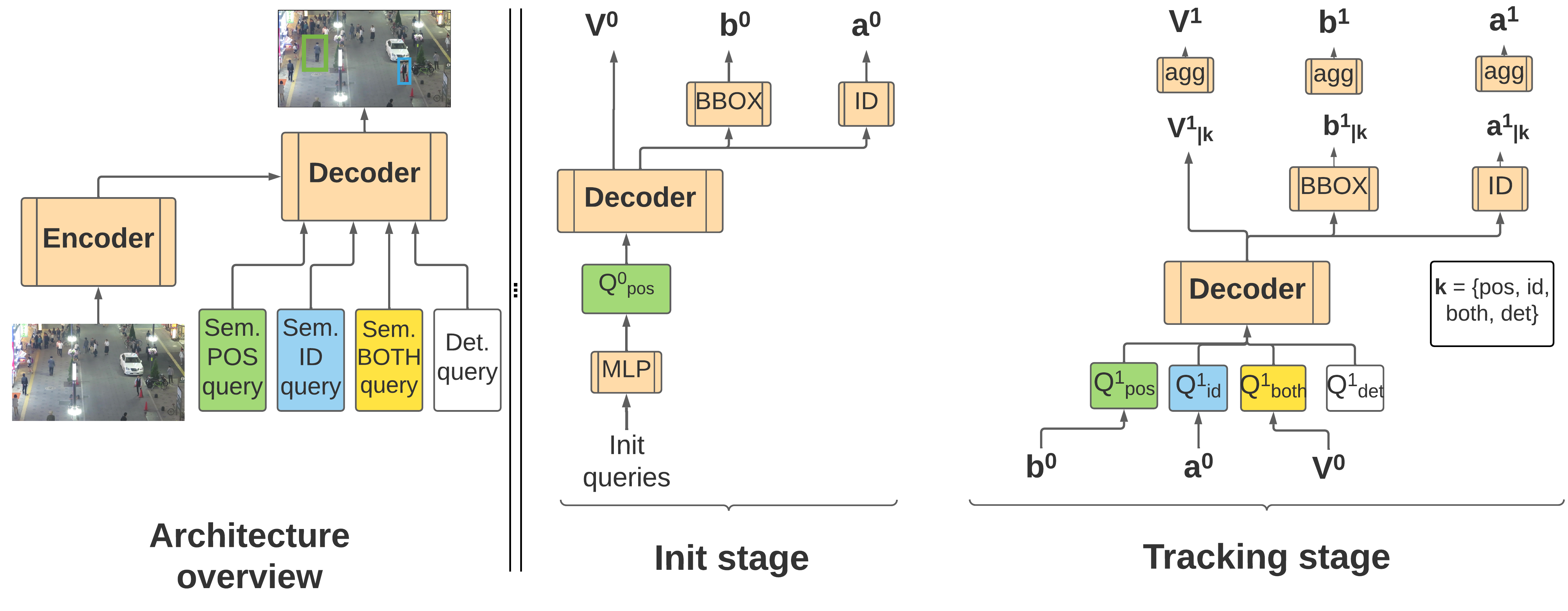}
\caption{\small We show the high level overview of MQT on the left. Two distinct training stages with a single query during initialization, and multiple queries during tracking stage are on the right.  Tracks are initialised either by using \texttt{det} (detection) queries, or with existing detections projected into semantic queries (e.g.\ $Q^0_{\texttt{pos}}$ as shown in the figure). Each query is then processed to obtain the decoder output ($V^0$), bounding-box prediction ($b^0$) and appearance vector ($a^0$). These are passed to the following frame in form of semantic queries, and their corresponding outputs ($V^1_{|k}$, $b^1_{|k}$, $a^1_{|k}$ respectively) are aggregated for each object to obtain final predictions ($V^1$, $b^1$, $a^1$).}
\label{fig:2stagetraining}
\end{figure}

The goal of multi-object tracking is to obtain the trajectories of $n$ objects over a sequence of frames from a video. For example, given the initial set of object locations (bounding boxes) in the first frame, the task is to predict a new set of bounding boxes and associate them to the correct objects for every subsequent frame thus forming trajectories. 

\noindent We formulate an auto-regressive tracking process as illustrated in Figure~\ref{fig:splash}.
At the current frame, the model
produces three outputs for each object: (1) a bounding box of the object's location in the current frame, (2) an appearance embedding of the visual appearance of the object given its location, and (3) a raw transformer-decoder embedding.
This information is then passed to the following frame in the form of semantic queries to the decoder. For the following frame, the model either looks for an object given its location, its appearance, or any additional information carried over by the raw decoder output in queries. The output embeddings are aggregated, and if the appearance output of an object at the frame $k$ matches the known appearance of the track (usually the appearance output at the frame $k-1$, but it can be earlier when using reID from memory to overcome occlusions), the location output is then added to the trajectory of an object. This makes our model applicable in generalised tracking scenarios where class information is not available. 

\noindent On a high level, this work is performed by a transformer~\cite{Vaswani17Transformers}. The current image is processed by a convolutional neural network and fed into a transformer-encoder, whereas all semantic queries from the previous image are fed into the transformer-decoder module -- see Figure~\ref{fig:2stagetraining}. This differs from traditional tracking-by-detection approaches (e.g.\ \citet{bergmann2019bells}) where detection is separate from data-association and where each step commonly uses separate embeddings. Our method merges these two steps into one, and the initial object embedding is disentangled within the transformer-decoder directly into embeddings used for detection and data-association (reID).

\noindent The rest of this section outlines the main parts of the model and their application for tracking. For more detailed information on architecture, hyperparameters and implementation details please refer to the supplementary material.

\subsection{Transformer-decoder queries} 

The key insight of our work is the fact that the queries passed to the decoder part of a transformer can be customized for the tracking task. For example, if we know the approximate bounding box of an object from a previous frame, then a query can be formed from this bounding box and used to search for the new position of the object in its vicinity
(in a similar manner to the RoI pooling module of a traditional two-stage detector that extracts the image embedding corresponding to the input bounding box, and is then used for bounding box regression and classification). 

\noindent What if we wanted to update the appearance (or maybe find the location of an object defined by its appearance)? We simply extend this approach by having a query encode the object appearance.

%
\noindent These \textit{semantic queries} are used in an auto-regessive manner for tracking, in that the output of the decoder of one frame is used as the query input for the subsequent frame.
We also include another type of query that is not used auto-regressively, but instead is applied independently on each frame. This second type of query, termed  (\texttt{det}), acts as spatial anchors in traditional detection transformers~\cite{carion2020DETR}, and allows the tracker to find new objects. For each image, a number of these queries are fed into the decoder module, and their outputs from the decoder feed into bounding-box and appearance heads. If any of these heads matches any recently ``lost" or ``current" track better than semantic queries, we assign the appearance and location outputs of that detection query to be a part of a trajectory.



\subsubsection{Location queries (\texttt{pos})}
Location queries are non-linear projections of an object bounding box, telling the model \textit{where} to look for an object. A bounding box is passed through a single-layer MLP that projects it to the model dimension $d$. It is then passed through the decoder producing a $d$-dimensional embedding $V_{|pos}$, which is then passed through: (a) a bounding-box regressor head trained to produce an output bounding box $b_{|pos}$, and (b) an appearance head trained to produce an output vector $a_{|pos}$ containing the information about object's appearance. 
The output of the bounding-box regressor head is used as a bounding box to form trajectories for an object.


\subsubsection{Appearance queries (\texttt{id})}
Sometimes objects get occluded, or their location can change due to camera shift or variable frame rate. In these cases, we need a mechanism that would tell the model \textit{what} to look for. To this end, we feed in the appearance query, which in our case is simply an output of the appearance head from a previous step $a^{i-1}$. This query is then also passed through the decoder to obtain vector $V_{|id}$ which is then passed through the bounding-box regressor and an appearance head producing outputs $b_{|id}$ and $a_{|id}$ respectively.

\noindent At tracking time, outputs of the appearance queries are used for track confirmation and re-identification (``matching" the track). Simply put, the track is only considered to be active if the cosine distance between outputs of the appearance query at the previous frame and current frame is smaller than a hyper-parameter $\tau_{\text{conf}}$, and the track is associated with $k$-th object if it is the object with smallest cosine distance to the output of the appearance query at the previous step. If a track became inactive (e.g.\ due to occlusions), we keep the track information for several frames in order to attempt to pick that track up again. For subsequent frames, we compare the appearance embeddings of detected objects to the ``known" inactive tracks in order to establish correspondence if the track is re-discovered. We empirically established that keeping several ($3$) frames worth of appearance information aids tracking performance overall.

\subsubsection{Joint queries (\texttt{both})}
Finally, we reason that a raw output embedding of the decoder $V$ contains valuable information about the track location and identity. Therefore, we form a joint query from $V^{i-1}$ that is simply a raw decoder output at the previous step. It is used to reinforce and strengthen the signal not captured from \texttt{pos} and \texttt{id} queries alone. They are passed through the same machinery as the formerly mentioned queries, producing outputs $V_{|both}$, $b_{|both}$ and $a_{|both}$. 

\subsubsection{Detection queries (\texttt{det})}
Much like anchor-queries in transformer-based 
detectors~\cite{carion2020DETR}, \texttt{det} queries are used detecting additional objects in tracking settings where additional detections are allowed. They are static, randomly initialized, trainable parameters used to detect and associate new objects to the existing semantic queries. They produce outputs $V_{|det}$ $b_{|det}$ and $a_{|det}$, which are then matched to the existing tracks independently from semantic queries.

\noindent For each frame, a number of
\texttt{det} queries are fed into the decoder module, and their outputs then pass through the bounding-box and appearance heads. If any of these heads matches any ``lost" or ``current" track better than semantic queries (often caused when the objects are occluded), we assign the appearance and location outputs of that detection query to be a part of a trajectory, and the outputs corresponding to that \texttt{det} query become semantic queries for the following frame.

\subsubsection{Query aggregation}
The eagle-eyed reader will have noticed that for every known object at a current frame $i$ there will be three sets of outputs $V_{|q}$, $b_{|q}$ and $a_{|q}$, for $q\in \{ \texttt{pos}, \texttt{id}, \texttt{both} \}$ (i.e.\ three outputs for each of the semantic query types). To make a final prediction for that object, we need to learn which one of the decoder outputs to trust. To this end we learn an aggregation function over each set of semantic queries, $V^{i}_{|q}$, $b^{i}_{|q}$ and $a^{i}_{|q}$, to obtain final results $V^{i}$, $b^{i}$ and $a^{i}$. Final appearance output for the frame will then be computed as 
\begin{align*} 
a^i &= \phi_q (a_{|q})   & q\in \{ \texttt{pos}, \texttt{id}, \texttt{both} \} 
\end{align*}
 where $\phi$ is an aggregation function. Each \texttt{det} query is passed through the aggregation function as well, but the effect is nullified as it is passed through it on its own.
In our work, we use collaborative-gating as an aggregaton function~\cite{liu2019use} as it performs better than other schemes (see ablation studies) such as taking an average of the embeddings.
In the case where different query types are missing (e.g.\ at initialisation or when dealing with \texttt{det} queries), we zero-pad the missing query types. To compensate for the implicit scaling introduced by missing query types. We follow~\cite{liu2019use} and remove the weights for missing queries, and then re-normalise the remaining weights such that they sum to one.

\subsection{Training the tracker} \label{sec:training_deets}
In order to be able to track objects frame-to-frame, we train the our model on two adjacent frames as illustrated in Figure~\ref{fig:2stagetraining}. To pre-train our model, we follow~\cite{trackformer} and simulate tracking data from COCO~\cite{COCO} (in order to run ablations on object detection and train the model for GMOT40~\cite{bai2021gmot40}) and Crowd-Human~\cite{shao2018crowdhuman}. As in~\cite{trackformer}, the adjacent frames are generated by applying random spatial augmentations of up to 5\% with respect to the original image size. For COCO~\cite{COCO}, we apply an additional constraint that no objects should be lost between the transformations.

\noindent At the initialization step, we compute bounding-box regression loss from either semantic or detection queries. In the tracking step, we optimize the model jointly for location prediction and appearance matching (i.e.\ tracking) for all objects initialized in the initialization step. 
We map the ground truth objects to the set of predictions from our model in one of two ways. For all known tracks we attempt to match them to the ground-truth based on the cosine similarity between the appearance vectors produced by the model in the initialisation and tracking step. If the track is ``lost" or undiscovered, i.e.\ the appearance vector at the current step doesn't match the appearance vector of the corresponding query at the initialization step (e.g.\ due to occlusions or objects moving out of the scene), we follow~\cite{carion2020DETR} and try to find injective minimum cost mapping between the ground truth and the set of predictions generated by detection (\text{det}) queries. In this way, we are able to train the model end-to-end, and the model is able to simultaneously reason about both known tracks and new, undiscovered objects. Note that during training, we always use both semantic and detection queries (100 per frame, following~\cite{carion2020DETR}). \\
\noindent \textbf{Class-agnostic vs class-specific tracking: }Modern detectors~\cite{girshick15fast, carion2020DETR} pass the region-of-interest embedding through regression and classification heads to obtain final bounding box and class predictions. When the object domain is known (i.e.\ we know the objects are people), the latter can very effectively be used in a tracking mode as a track confirmation mechanism -- if an object is a pedestrian, mark it as a part of a track. While this approach is intuitive, it has several drawbacks. The first and the most obvious one is the fact that the object class has to be known in advance, in order to finetune the classification head for that particular object class. Additionally, it does not carry information about the particular instance, thus necessitating the use of some other data-association mechanism. 
When our model is trained for tracking, we ignore the classification head and instead determine the track validity and identity by comparing the cosine distance between two subsequent appearance outputs and picking a minimum one. While this has a slight detrimental impact on our overall tracking performance (see ablation studies), it allows us to apply our tracker in a class-agnostic fashion on generalised-MOT task, and we empirically found that our model retains track identities better.\\
\noindent \textbf{Loss functions: } The final transformer set-prediction loss~\cite{carion2020DETR} at each tracking step is computed over all semantic and detection queries, in the same fashion as~\cite{trackformer}. However, we use the proposed appearance loss instead of the class-prediction loss of~\cite{trackformer}. The appearance loss is formally defined in Sec~1.4 of the supplementary material.

\subsection{Enhancing our tracking results}
\noindent \textbf{Memory: }One important way of enhancing the performance of a tracker is to increase the memory of a track. Specifically, for each \textit{known} track, we are keeping the information of the last frame in which the track was active. This makes sense as the appearance is unlikely going to change in such a short time span. However, there can be instances of occlusions, rotations, and tracks becoming inactive for (relatively) long periods of time where having more data points about the appearance of the object could be useful. Hence we explore keeping multiple frames worth of appearance information. We found that keeping the appearance information of the first and the last five frames gives us optimal results. We simply proceed to use the minimum distance between the current appearance vector and the ``memory" appearance vectors.

\noindent \textbf{Multi-hypothesis tracking: }Transformer-based methods such as ours tend to suffer from a large number of false-positives as multiple query-anchors can detect the same object, an effect we largely mitigate by considering only objects that have maximum appearance similarity to the track at the previous frame. This can cause more identity shifts as in the ambiguous cases, model might shift between two tracks (which can lead to it losing the track completely). We relax this requirement and keep track of the top-$k$ possible track-candidates (that are still within the threshold $\tau_{\text{conf}}$), and then choose the longest tracked sequence as a part of our trajectory.

\noindent \textbf{Leveraging multiple queries: } Keeping multiple hypothesis for every object however can significantly increase the memory requirement, especially when keeping multiple hypothesis is not always necessary. To this end, we leverage our multi-query setup and introduce a simple heuristic in order to reduce the memory requirement in the cases where the model is completely certain in an existing track. Specifically, we define confidence as an agreement between two query types: if the distance between the appearance vectors corresponding to the \text{pos} and \text{id} queries before aggregation is smaller than a hyper-parameter $\tau_{\text{agree}}$, we do not consider additional hypothesis.

\section{Experiments}

In this section, we first present the results of our best model on various tracking tasks. Then, we present the ablation study in which we experimentally verify the design choices from the previous section. Final hyper-parameters for each dataset are given in the supplementary material.

\subsection{Datasets and tasks}
\label{sec:datasets}

\noindent\textbf{Person-MOT:} 
The \textbf{MOT17} dataset consists of train and test sets, each with 7 sequences containing full-body bounding boxes of pedestrians for up to 51 people in a sequence. Three sets of public detections are provided: DPM, Faster R-CNN and SDP~\cite{Felzensqalb2010DPM, ren2015faster, yang2016SDPnet}. For our tracking ablation studies, we select a single sequence from the training split as a validation set, following~\cite{bergmann2019bells}.\\
Compared to MOT17, \textbf{MOT20}~\cite{MOT20} is more challenging as it contains much more crowded scenes, some with up to 220 pedestrians.
Finally, we evaluate our model on the ``person" class of the \textbf{TAO}~\cite{achal2020TAO} dataset. See below for more details about the dataset.
\\
\noindent\textbf{Generalized-MOT: } MOT challenges focus on a specific object category
of interest (pedestrians) and rely on models trained specifically to recognise them. In contrast, generalised MOT (GMOT) requires no prior knowledge of the objects to be tracked, and is an evaluation of class-agnostic tracking.  GMOT-40 contains 40 sequences which cover ten categories loosely related to categories of popular object detection datasets, with four sequences per category. Each sequence contains multiple objects of the same category.
\\
\noindent\textbf{Tracking any object: } TAO~\cite{achal2020TAO} is a diverse dataset for a task of tracking any object. It
consists of 2,907 high resolution videos, captured in diverse environments, with a vocaulary of over 800 objects. The dataset is split into ``train", ``validation", and ``test" splits, containing 500, 988 and 1,419 videos respectively. 
\\
\noindent \textbf{Metrics: }Aspects of MOT are evaluated with different standardised metrics~\cite{bernardin2008MOTmetrics}. The community focuses on two complementary metrics: Multiple Object Tracking Accuracy (MOTA) which focuses on track coverage of the detections, and Identity F1 Score (IDF1) which focuses on identity preservation across the tracks~\cite{ristani2016idf1metric}. For TAO, we compute mAP metric using 3D IoU (with a default threshold of 0.5) as specified in~\cite{achal2020TAO}.

\noindent \textbf{Private vs.\ public tracking:} For the evaluation on MOT datasets such as MOT17~\cite{MOT16and17} tracking works often refer to the \textit{private} and \textit{public} detections. \textit{Public} detections are provided with the dataset and allow for a comparison of tracking methods independent of the object detection performance of the model. \textit{Private} detection setting during evaluation allows for the use of detections obtained in any other way. For our evaluation, if the setting is marked as \textit{public}, we strictly use detections provided with the dataset; if the setting is marked \textit{private}, we initialise the tracks with detection \texttt{det} queries only and at each step we allow a number of them to detect potentially missed objects.
In addition to these traditional settings, we denote two additional ones. If setting is marked as \textit{private \& public}, we use the detections provided \textit{and} \texttt{det} queries, and rely on the set-matching algorithm to find an optimal mapping to the ground-truth tracks. If any of the settings is marked with ``private+", it means we augment our detections with those obtained from a state-of-the-art object detection model~\cite{yolox2021} trained separately and feed them into the decoder as additional \texttt{pos} queries. For a more detailed description of each stage in the context of our model, please refer to the supplementary material. 

\begin{table}[!tbp]
\small
\centering
    \caption{\small Comparison of modern MOT methods evaluated on MOT17 and MOT20 test sets. ``+" in our private setting denotes use of externally computed private detections from~\cite{yolox2021}. Models denoted with ``*" are not associated with a publication. For fairer comparison, we specifically mark models with in-model association solvers (IAS).}
    \label{table:mot}
    \begin{tabular}{@{}l|cccccc@{}}
        \toprule
        \multicolumn{1}{l}{\multirow{2}{*}{Method}} & \multirow{2}{*}{Setting} &
        \multirow{2}{*}{IAS} & 
        \multicolumn{2}{c}{MOT17} & \multicolumn{2}{c}{MOT20} \\ \cmidrule(l){4-7} 
        \multicolumn{1}{l}{} & &  & MOTA~$\uparrow$ & IDF1~$\uparrow$ & MOTA~$\uparrow$ & IDF1~$\uparrow$ \\ \midrule
        \midrule
        Tracktor++~\cite{bergmann2019bells} & Public & & 56.3 & 55.1 & 52.6 & 52.7 \\
        CenterTrack~\cite{zhou2020centertrack} & Public & & 60.5 & 55.7 &  &  \\
        *Trackformer~\cite{trackformer} & Public & \checkmark & 62.5 & 60.7 &  &  \\
        *TransCenter~\cite{xu2021transcenter} & Public & & \textbf{71.9} & 62.3 & 62.3 & 50.3 \\
        SiamMOT~\cite{shuai2021siammot} & Public & & 65.9 & 63.1 & &  \\
        MQT (ours) & Public & \checkmark & 65.4 & \textbf{63.3} & \textbf{63.3} & \textbf{55.8} \\ \midrule
        CenterTrack~\cite{zhou2020centertrack}  & Private & & 67.8 & 64.7 &  &  \\
        *Trackformer~\cite{trackformer} & Private & \checkmark & 65.0 & 63.9 &  &  \\
        *TransCenter~\cite{xu2021transcenter} & Private & & 73.2 & 62.2 & 61.9 & 50.4 \\
        *RelationTrack~\cite{yu2021relationtrack}& Private & & \textbf{73.8} & \textbf{74.7} & \textbf{67.7} & \textbf{70.5} \\
        MeMOT~\cite{cai2022memot}& Private & \checkmark & 72.5 & 69.0 & 63.7 & 66.1 \\
        MQT (ours) & Private & \checkmark & 66.5 & 65.2 & 62.1 & 62.5 \\
        MQT (ours) & Private+ & \checkmark & 68.2 & 65.8 & 66.1 & 64.5 \\ 
        \midrule
        MQT (ours) & Prv\&Pub & \checkmark & 67.9 & 64.2 & 64.8 & 63.9 \\
        MQT (ours) & Prv+\&Pub & \checkmark & 69.4 & 65.9 & 66.9 & 65.7 \\
        \bottomrule
    \end{tabular}
\end{table}

\smallbreak

\subsection{Results}
\noindent\textbf{MOT17: } We evaluate our model on the MOT17~\cite{MOT16and17} test set and report the results in Table~\ref{table:mot}. For private detections, our results are comparable to the most modern works even though our model is not inherently trained for detection. 
When detection queries are added to the initialization stage (technically making our detection not truly public, but not fully private either as we do leverage the detections provided with the dataset and augment them with better detections), it gains additional $3.2$ MOTA points advantage over fully public setting, and it even surpasses most trackers using private detections only.

\noindent\textbf{MOT20: }We also evaluate our model on a much more challenging MOT20 dataset. Results can be seen in Table~\ref{table:mot}.
Our performance is on par with modern trackers in public detection setting, however, our model suffers in a private setting. This is largely due to the superior performance the two-stage detectors exhibit on frames with many small objects in them. When we make use of the superior detections, our results are much more competitive. 

\noindent It is worth noting that RelationTrack~\cite{yu2021relationtrack} and TransCenter~\cite{xu2021transcenter}, methods that significantly outperform ours, separate the detection part of the model completely from the data-association part whilst still training end-to-end. TransCenter, however, is limited by the detection capabilities of their backbone, which is what allows us to outperform their model on the more challenging MOT20 dataset~\cite{xu2021transcenter}. The RelationTrack algorithm fully separates the detection and association, and refines generated tracklets in a two stage procedure at inference time~\cite{yu2021relationtrack}. In the supplementary material we show that a similar method can aid our model as well with a trade-off in inference speed and flexibility.


\noindent\textbf{TAO-person: } Finally, we evaluate our model on the person category of the TAO dataset~\cite{achal2020TAO}. We tune the threshold parameters on a small hold-out section of the training set, but we do not re-train our model on TAO. We include our re-implementation of~\cite{bergmann2019bells} and~\cite{trackformer} evaluated following the same protocol and using the same set of detections whenever possible for a fair comparison. Full results can be seen in Table~\ref{tbl:tao}. MQT outperforms the reported performance of~\cite{bergmann2019bells} by $5.3$ MOTA points, and improves upon our re-implementation of~\cite{bergmann2019bells} and~\cite{trackformer}.

\begin{table}[t]
\small
\centering
\caption{\small Evaluation of state-of-the-art person trackers on the person category of the TAO dataset. Methods marked with `*' denote our re-implementation of the methods and the evaluation protocol. Tractor++ and \textit{MQT} are using the same private detections whilst Trackformer uses its own detections.}
\label{tbl:tao}
        \begin{tabular}{@{}l|cc@{}}
        \toprule
        Method & MOTA~$\uparrow$ & IDF1~$\uparrow$ \\ \midrule
        Tractor++~\cite{bergmann2019bells} & 66.6 & 64.8 \\
        *Tractor++~\cite{bergmann2019bells} & 68.1 & 66.1 \\
        *Trackformer~\cite{trackformer} & 71.3 & 67.2 \\
        MQT (ours) & \textbf{71.9} & \textbf{69.6} \\ \bottomrule
    \end{tabular}
\end{table}

\subsection{Expanding tracking capabilities: }
Our model formulation allows us to successfully apply our model ``as-is" to more general tracking scenarios such as \textit{class-agnostic} MOT or \textit{tracking any object}. To demonstrate this capability, we evaluate our model on \textbf{GMOT40}~\cite{bai2021gmot40} and \textbf{TAO}~\cite{achal2020TAO} datasets. 

\noindent\textbf{GMOT40: } MOT17 and MOT20 datasets contain multiple instances of pedestrians on the street. Intuitively, this discards a lot of detection capacity of models trained on detection data and assumes prior knowledge of the domain. To this end, we evaluate our model on GMOT40 -- a generalised MOT benchmark where categories are related (but not completely overlapping) with common detection datasets. Our model outperforms all baseline benchmarks on the test set, largely due to its class-agnostic nature. For this task, we forgo CrowdHuman~\cite{shao2018crowdhuman} pre-training and train the models on tracking data simulated from COCO~\cite{COCO}.

\noindent\textbf{TAO: }To demonstrate the promising generalization of our MQT model even for the task outside of the ``traditional" MOT scope, we present results on the \textit{val} split TAO dataset~\cite{achal2020TAO}. First three rows of Table~\ref{table:tao} present ``user-initialised" tracking setting with oracle ground truth class assigned using the protocol described in~\cite{achal2020TAO}. Detections are either provided (MaskRCNN~\cite{he2017mask}, akin to ``public" setting on MOT tasks), or directly computed by us (``private" setting). Note that a lack of fully trained detector hurts our performance on this task, as can be seen by a boost seen when additional detections are provided to our model.

\begin{table}[]
\hfill
\parbox[t]{.45\linewidth}{

    \small
    \caption{\small Performance of our tracker on one-shot GMOT protocol, as described by~\citet{bai2021gmot40}}
    \label{table:gmot40}

    \begin{tabular}{@{}l|cccc@{}}
        \toprule
        Method & MOTA~$\uparrow$ & IDF1~$\uparrow$ & MT~$\uparrow$ & ML~$\uparrow$ \\ \midrule
        MDP & 19.80 & 31.30 & 142 & 1161 \\
        FAMNet & 18.00 & 28.30 & 166 & 1197 \\
        Ours & \textbf{23.95} & \textbf{31.06} & \textbf{182} & \textbf{1077} \\ \bottomrule
    \end{tabular}
}
\hfill
\parbox[t]{.5\linewidth}{
    \small
    \caption{\small User-initialized tracking results on ``val" split of the TAO dataset.}
    \label{table:tao}
    \begin{tabular}{@{}lcc@{}}
        \toprule
        Tracking     & Detection & Track mAP \\ \midrule
        SORT~\cite{achal2020TAO}  & MaskRCNN~\cite{he2017mask} & 30.23     \\
        Ours    &  ours  & 32.11     \\ 
        Ours    &  \begin{tabular}[c]{@{}c@{}}ours $+$\\ MaskRCNN~\cite{he2017mask}\end{tabular} & \textbf{39.62 } \\
        \bottomrule
    \end{tabular}
}
\end{table}


\subsection{Ablation studies}
In this section, we validate the model design choices of our tracker outlined in Section~\ref{sec:method}. For additional ablations, please refer to our supplementary material.

\subsubsection{Tracking}

\begin{table*}[!htb]
\centering
\caption{Ablation study of \textit{MQT} on MOT17 held out training sequence.}
\label{tbl:abl_track}

\scriptsize
    \centering
    \begin{subtable}[t]{.3\linewidth}
        \begin{tabular}{@{}llllll@{}}
        \toprule
        \texttt{det} & \texttt{id} & \texttt{pos} & \texttt{both} & \multicolumn{1}{c}{MOTA} & \multicolumn{1}{c}{IDF1} \\ \midrule
        \checkmark & \checkmark & \checkmark & \checkmark & 68.3 & 66.1 \\
        \checkmark &  &  &  & 55.9 & 54.7 \\
        & \checkmark &  &  & 52.7 & 54.1 \\
        &  & \checkmark &  & 56.1 & 55.8 \\
        &  &  & \checkmark & 56.9 & 56.2 \\ \bottomrule
        \end{tabular}
        \caption{\scriptsize Single vs multi-query tracking performance. For a full table with all permutations, please refer to the supplementary material.}
    \end{subtable}
    \hspace{4pt}
    \centering
    \begin{subtable}[t]{.3\linewidth}
     \centering
    \begin{tabular}{@{}l|cc@{}}
        \toprule
        Method & MOTA & IDF1 \\ \midrule
        Heuristic & 65.7 & 64.6 \\
        Avg. pool & 65.8 & 64.7 \\
        Max. pool & 61.4 & 59.3 \\
        \begin{tabular}[c]{@{}c@{}}Colaborative\\ gating\end{tabular} & 68.3 & 66.1 \\ \bottomrule
        \end{tabular}
        \caption{\scriptsize Comparison of various query aggregation methods for tracking, evaluated on held-out validation sequence from MOT17.}
    \end{subtable}
    \hspace{4pt}
    \begin{subtable}[t]{.3\linewidth}
    \centering
        \begin{tabular}{@{}ccll@{}}
            \toprule
            \begin{tabular}[c]{@{}c@{}}Class \\ head\end{tabular} & \begin{tabular}[c]{@{}c@{}}Track \\ confirmation\end{tabular} & MOTA & IDF \\ \midrule
            yes & class & 68.5 & 66.0 \\
            yes & appearance & 68.2 & 66.1 \\
            no & appearance & 68.3 & 66.1 \\ \bottomrule
        \end{tabular}
    \caption{\scriptsize MQT trained with and without classification head. Results indicate that training the classification head is not necessary from the tracking standpoint when appearance query is used for track-confirmation.}
    \end{subtable}
\end{table*}

We examine the impact of  various tracking design choices on a single held-out sequence of MOT17, following the ablation procedure described in~\cite{bergmann2019bells}.

\noindent\textbf{Single- vs.\ multi-query decoder: }We investigate the gain in having multiple types of semantic queries (rather than just having \texttt{pos} only for example). The results of single-vs-multi query tracking performance are given in Table~\ref{tbl:abl_track}~(a), whilst full results can be found in the supplementary material. When only \texttt{det} queries are used, we get the performance akin to what we would get with a detection transformer only. Having an auto-regressive component (e.g.\ \texttt{both} query) improves performance by a small margin (1 MOTA point). The performance benefit of multiple queries is clear, outperforming any single-query tracker by 11 MOTA points.

\noindent\textbf{Feature aggregation -- tracking: }We examine different ways to aggregate outputs with respect to multiple decoder query-types. The \textit{heuristic} method refers to simply using the location output w.r.t.\ the previous location to regress the bounding box, and appearance output w.r.t.\ the previous appearance for track-confirmation and reID (full evaluation of heuristic method, including all permutations of three query-types can be found in the supplementary material). We found that collaborative gating~\cite{liu2019use} performs better than any other aggregation method by $2.5$ MOTA points.  Full results can be seen in Table~\ref{tbl:abl_track}~(b).

\begin{table}[!htb]
\centering
    \caption{Various methods of improving our tracker's performance.}
    \label{tbl:abl_perfupgrade}
\centering
\scriptsize
    \centering
    \begin{subtable}{.48\linewidth}
    \centering
    \begin{tabular}{@{}lc|cc@{}}
        \toprule
        \# frames & \multicolumn{1}{c}{\begin{tabular}[c]{@{}c@{}}dist \\ metric\end{tabular}} & MOTA & IDF1 \\ \midrule
        1 (F) & n/a & 67.1 & 65.8 \\
        1 (L) & n/a & 68.3 & 66.1 \\
        2 (F + L) & avg & 69.2 & 66.3 \\
        2 (L) & avg & 69.0 & 66.1 \\
        3 (F + L2) & avg & 69.6 & 66.5 \\
        6 (F + L5) & avg & 69.6 & 66.7 \\
        6 (F + L5) & min & 70.0 & 66.9 \\ \bottomrule
        \end{tabular}
        \caption{\scriptsize
Impact of various memory sizes (and metric for computing the appearance distance) on tracking performance, evaluated on held-out validation sequence from MOT17. (F) indicates the first frame, (L$k$) indicates the last $k$ frames.}
    \end{subtable}
    \hspace{4pt}
    \begin{subtable}{.48\linewidth}
    \centering
        \begin{tabular}{@{}lcc@{}}
        \toprule
        \begin{tabular}[c]{@{}l@{}}Number of \\ proposals\end{tabular} & MOTA & IDF1 \\ \midrule
        1  & 68.3 & 66.1 \\
        3  & 69.7 & 67.8 \\
        5  & 70.4 & 68.4 \\
        5 (MQC)& 70.5 & 68.6 \\
        10  & 69.5 & 68.1 \\ \bottomrule
        \end{tabular}
        \caption{\scriptsize
Multi-hypothesis tracking: analysing the impact of keeping multiple candidates for each track. Note that multi-query confirmation (MQC) does not impact the results, but it reduces the computational requirement by only keeping multiple tracks when model confidence drops.}
    \end{subtable}


\end{table}

\noindent\textbf{Class-agnostic vs.\ class-specific tracking: }The majority of modern trackers that fall into tracking-by-detection category rely on class-specific information in order to achieve good performance on MOT challenges. We ask if that is really necessary, given that our model is capable of producing appearance-specific embeddings for every track. To this end, we compare two different track confirmation methods: the traditional approach which uses a class-confidence threshold to confirm the track, and our suggested approach that relies on setting a threshold for the distance between the appearance information of two subsequent frames. In order to rule-out the effect of training the classification head on downstream performance we also show results when our scheme is used, but classification head is still being trained. Using classification score for track-confirmation is marginally more effective (by $0.3$ MOTA), but we find that the performance benefit is not worth the limitations it poses (mainly inability to track ``unknown" classes). Furthermore, we find that training the classification head during tracking pre-training has little to no impact on downstream tracking performance if the classification score is not being used. Full results are given in Table~\ref{tbl:abl_track}~(c).\\

\noindent\textbf{Appearance memory size: } In Table~\ref{tbl:abl_perfupgrade}~(a), we explore the impact of various memory sizes (and metric for computing the appearance distance) on tracking performance. We either compare the distance of a current track to the minimum distance of all the embeddings (min), or to the average-pooled embedding (avg). As a broad trend, the more memory we store, the better the performance get. Due to computational constrants, we were not able to extend this beyond six frames, however, even between 3 and 6 frames the performance difference becomes marginal, indicating we might hit diminishing returns.\\

\noindent\textbf{Multi-hypothesis tracking: } In order to evaluate the performance benefits of multi-hypothesis tracking, we evaluate our model with a varying number of proposals at each step. The tracking performance saturates at 5 proposals, indicating that there is no benefit in keeping more than this. Furthermore, using the distance between \texttt{loc} and \texttt{id} queries as a measure of confidence in predictions of known objects doesn't impact performance of the model whilst reducing memory requirements. Full results are given in Table~\ref{tbl:abl_perfupgrade}~(b).

\section{Conclusion} 

We have introduced the multi-query transformer tracking model that achieves admirable performance on several known-class multi-object tracking challenges, while simultaneously outperforming all baselines on class-agnostic generalised multi-object tracking benchmark. We show the benefit of using decoupled \textit{semantic} decoder queries for both object detection and tracking, and we conjecture that similar strategy can be employed in different areas of computer vision. 

\clearpage

\setcounter{page}{1}
\appendix
\section{Implementation details} \label{sec:1}
In this section, we describe the details of the model, and training recipes of our tracker. Code, models and training configurations will be made publicly available upon publication.

\subsection{Detection transformer} \label{sec:11}
Our detector transformer is based on deformable-DETR~\cite{zhu2021ddetr} transformer with ResNet50~\cite{he2016deep} backbone. We use a four feature-level deformable detection module~\cite{zhu2021ddetr}, 256 dimensional embeddings and 2048 feed-forward dimension size, 6 encoder layers and 6 decoder layers with 8 attention heads. No additional modifications to the transformer architecture were made.

\subsection{Training details}\label{sec:12}
All our models are initialised from an object detector~\cite{zhu2021ddetr} that is pre-trained on the COCO dataset with additional \texttt{pos} queries (that encode perturbed ground-truth bounding box). Encoding perturbed ground-truth bounding boxes and passing them as \texttt{pos} queries during detection training not only boosts detection performance (as seen in Table~\ref{tbl:cocoabl}), but reduces the number of pre-training epochs before convergence. Since this is detection-only training, we do not train with queries which are used over multiple frames (\texttt{both} and \texttt{id}).

\noindent \textbf{MOT17/20:} For MOT challenges, the model is first trained on simulated motion pairs of images from the CrowdHuman dataset for 50 epochs with backbone learning rate of $1\mathrm{e}-5$, and encoder-decoder learning rate of $1\mathrm{e}-4$, and we reduce the learning rate by a factor of 10 after the 40th epoch. Then, the model is finetuned for the particular downstream MOT dataset (e.g.\ MOT17) for an additional 20 epochs, reducing the learning rate again half-way through the finetuning.

\noindent \textbf{GMOT40: }For generalised MOT, we train the model on pairs of images from COCO with motion simulated by an affine transformation (as described in the main body of the paper) for 50 epochs with the same initial learning rate as above, decreasing the learning rate after 20th and 40th epoch.

\noindent \textbf{TAO: } For TAO, we follow the same training procedure as for GMOT40, but additionally fine-tune the model on TAO training set for an additional 10 epochs with learning rate of $1\mathrm{e}-5$ across all modules.  For the TAO-person dataset, we use the model trained on MOT17, and tune the $\tau_{\text{conf}}$ parameter on the TAO-person training set.

\subsection{Tracking hyper-parameters} \label{sec:13}
In this subsection we give the details on two hyper-parameters of \textit{MQT} model: $\tau_{\text{conf}}$ and $\tau_{\text{agree}}$. $\tau_{\text{conf}}$ is a track-confirmation hyper-parameter. A track is only considered active if the similarity of the appearance query corresponding to the object $k$ at frame $i-1$ and $i$ is greater than $\tau_{\text{conf}}$.  $\tau_{\text{agree}}$ is used in a version of our model where we leverage multiple queries in order to determine track quality. If the cosine distance between the appearance vectors corresponding to the \text{pos} and \text{id} queries refering to the same object $k$ (before aggregation) is smaller than a hyper-parameter $\tau_{\text{agree}}$, we do not consider additional hypothesis. On datasets where the model is fine-tuned, hyper-parameters are tuned via linear search on the held-out validation set. For datasets where it would be too expensive to fine-tune the model (e.g.\ TAO), we determine the optimal parameter by conducting a linear search with a pre-trained model. Specifically, we run a 5-fold cross-validation on a held-out validation set. The values are presented in Table~\ref{tbl:hyper}.

\begin{table}[]
\centering
\caption{Additional model hyper-parameter values. \\}

\begin{tabular}{lcc}
\toprule
Dataset & $\tau_{\text{conf}}$ & $\tau_{\text{agree}}$  \\ \midrule
MOT17 & 0.75 & 0.1 \\
MOT20 & 0.65 & 0.05 \\
TAO-person & 0.80 & 0.2\\
GMOT40 & 0.85 & 0.2 \\
TAO & 0.65 & 0.1 \\
\bottomrule
\end{tabular}
\label{tbl:hyper}
\end{table}

\subsection{Appearance loss} \label{sec:14}
For tracking purposes, we train the model with an appearance head on top of the transformer decoder. The purpose of the appearance head is to make two corresponding instances similar in the latent space, and at the same time push them away from all other instances of the same class (or indeed from any non-tracked objects). For this to be satisfied, the appearance head must remove the location information from the raw embedding that is outputted from the decoder. 

Consider object $q^k \in Q$ being the appearance encoding of the object $q$ at frame $k$, where $Q$ is a set of all outputs from the decoder for that frame. For simplicity of notation, let us denote a set of all {\em other} outputs from the decoder in the same frame as $Q_{-} = Q \setminus q$. We then compute the appearance loss for $q$ as
$$\mathcal{L}_q = -log\frac{\text{exp}(q^{k+1} \cdot q^k)}{\sum_{j\in Q_{-}}\text{exp}(q^{k+1}  \cdot j^k)}$$
In this way, we can train the appearance head to associate related objects without any additional supervision.

Note, however, that the number of negative samples outweighs the positives. During training with \texttt{det} queries, the size of $Q_{-}$ gets large ($>100$, depending on the number of \texttt{det} queries). To offset this, we down-weight the negative samples by a factor of $0.1$. 

Note that the appearance head replaces what would be a class-prediction head in the traditional object-detection model, and appearance loss replaces the classification loss in detection-transformer framework. Therefore, we need to assign it the matching costs for Hungarian matching algorithm and loss coefficients in order for the transformer to train and utilise the new information~\cite{carion2020DETR}. We cross validate these hyper-parameters on the MOT17 dataset, and train the final models with matching cost of $1$ and loss coefficient equal to $2$.

\section{Tracking protocols} \label{sec:2}

\begin{figure}
     \centering
     \caption{Various tracking settings are illustrated bellow. We show different semantic queries (\texttt{pos}, \texttt{id}, \texttt{both}) in colour (green, blue and yellow respetively), and static \texttt{det} queries are shown in white. Best seen in colour.}
     \label{fig:tracking}
     \begin{subfigure}[b]{0.48\textwidth}
         \centering
         \includegraphics[width=\textwidth]{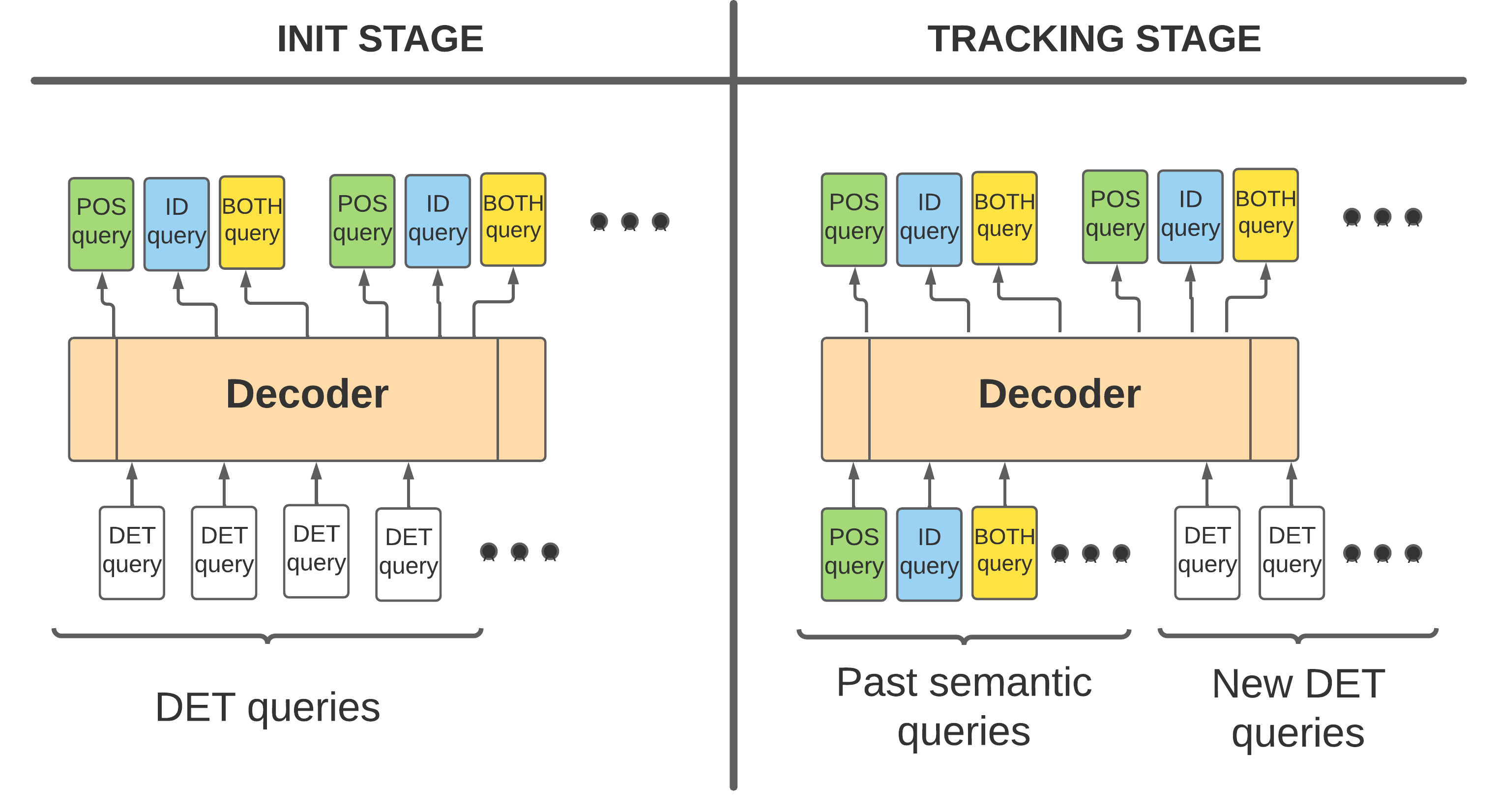}
         \caption{Illustration of initialisation and tracking for \textit{private} detection setting. \texttt{det} queries are a set of learnt vectors that act as spatial anchors and are used for detection at each step.}
     \end{subfigure}
     \hfill
     \begin{subfigure}[b]{0.48\textwidth}
         \centering
         \includegraphics[width=\textwidth]{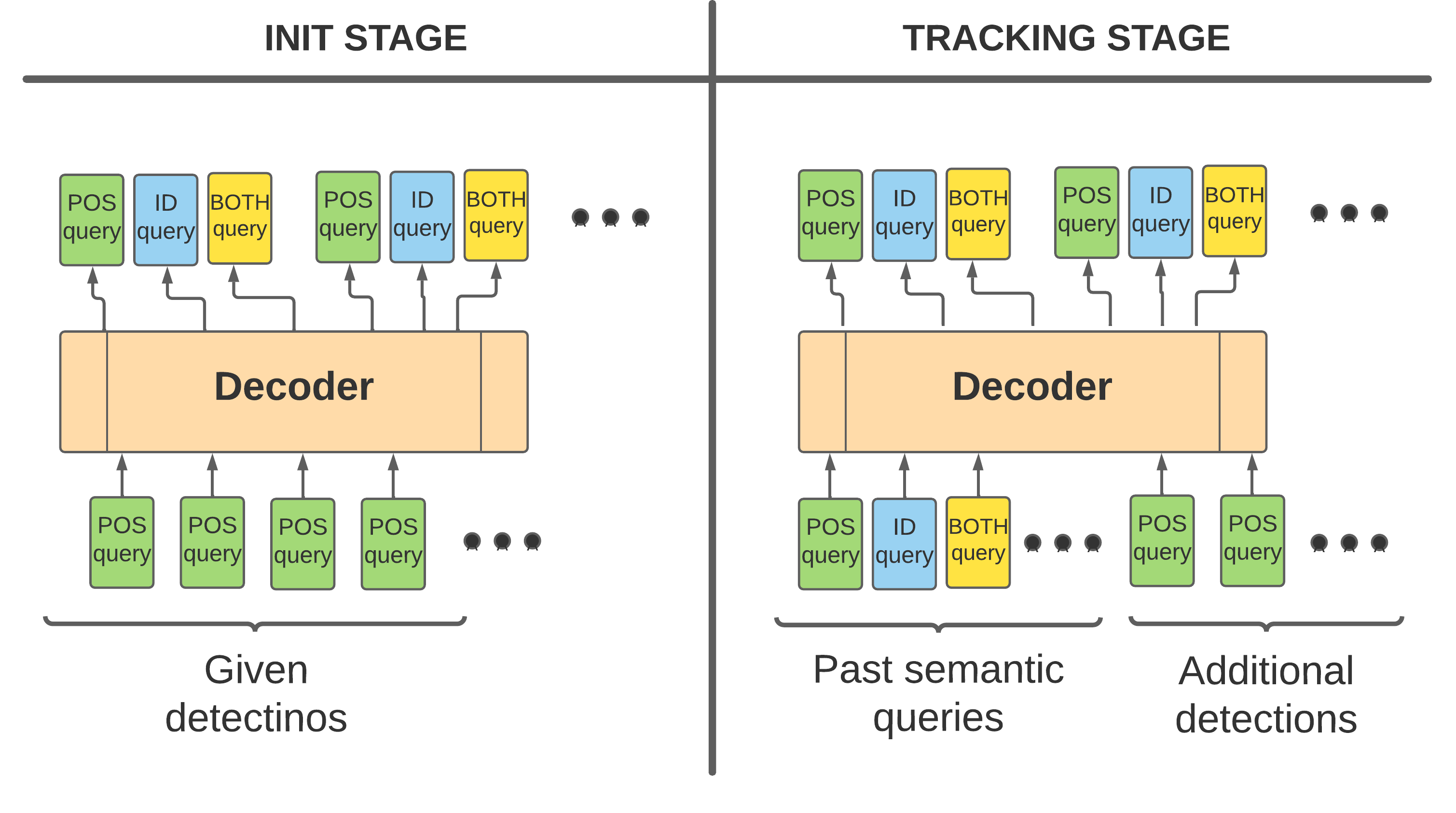}
         \caption{Illustration of initialisation and tracking for \textit{public} detection setting. Detections supplied with the sequences are passed to the model as \texttt{pos} queries.}
     \end{subfigure}

     \begin{subfigure}[b]{0.8\textwidth}
         \centering
         \includegraphics[width=\textwidth]{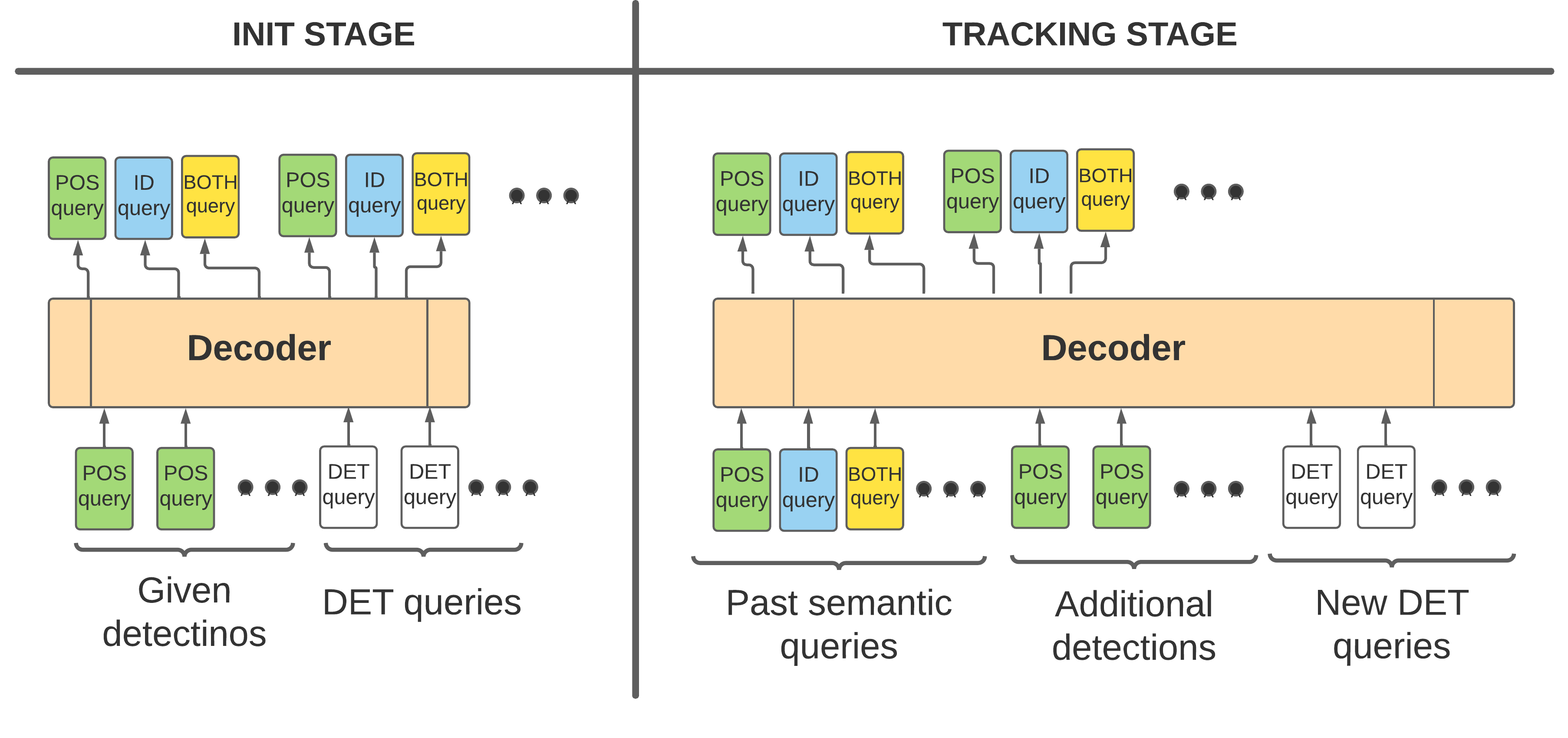}
         \caption{Illustration of initialisation and tracking for \textit{private}~\&~\textit{public} detection setting.}
     \end{subfigure}
        
\end{figure}

In this section we expand on the use of varying tracking protocols, and how they reflect on our model. Despite the intuitive simplicity of our model, the various intricacies of tracking benchmarks require different types of inputs to the decoder. Therefore, we describe various scenarios for each dataset. Visual illustration of the protocols can be seen in Figure~\ref{fig:tracking}.

\subsection{MOT} \label{sec:21}

Fundamentally, MOT tasks fall in one of two categories: private and public. In the \textit{private} setting, the task of the model is to detect all possible objects of interest in each frame, and track them thus forming trajectories over multiple frames. In the \textit{public} setting, we are given detections for each frame. The sole task of the model is then to possibly refine and associate the detections with one another thus forming trajectories.

\noindent  \textbf{Public: }
In the public setting, the detections are given. In that case, we initialise the sequence with the given detections, passed to the transformer decoder as semantic \texttt{pos} queries. At the tracking step, we propagate semantic queries for all tracked objects, and add on the detections for that particular frame as additional \texttt{pos} queries.

\noindent \textbf{Private: }In the private setting, we initialise the tracking sequence with a number of static \texttt{det} queries. They are a set of learnt vectors that are fed into the decoder at each step, and don't change between frames (hence static). If the appearance output of any two \texttt{det} queries have a similarity greater than $\tau_{\text{conf}}$, we establish the two respective queries as known object tracks. The three semantic queries for each object track are then passed to the following frame in an auto-regressive manner, together with additional \texttt{det} tracks. This initialisation approach is  sub-optimal compared to the class-specific models that output confidence score per track as class probabilities (though this is only applicable in class-based tracking by detection). However, empirically, we notice that the difference between the two methods is minor (see Table~5~(c) of the main paper).

\noindent  \textbf{Private \& Public: }
In a setting we denote as ``Private \& Public", we combine ``private" and ``public" setting. Namely at both initialisation and tracking stages feed in both \texttt{det} queries as well as the detections passed in as \texttt{pos} semantic queries.

\noindent \textbf{Private+: } In a setting we denote as ``Private$+$", we follow the same procedure as in ``Public", but replace the given detections with independently obtained detections from a state-of-the-art detector~\cite{yolox2021}.


\subsection{TAO} \label{sec:22}

We report the results on TAO in a ``user initialised" setting with a standard ``init first" approach~\cite{achal2020TAO}. For each object in TAO, the tracker is initialised using the fist frame an object appears in and it runs for the rest of the video. Similar to their experiments, we consider the object absent when the confirmation threshold falls under a certain value ($0.65$, cross validated on TAO training set). Finally, the tracks are supplied with a class oracle in order to be able to report the ``mAP" score, as in~\cite{achal2020TAO}.

\section{Additional ablation studies} \label{sec:3}
In this section, we expand upon the ablation studies in the main body of the paper (Sec.~4.3). First, we verify the effectiveness of our semantic queries in a simpler object-detection scenario. Then we investigate the impact each query type has on tracking to further expand our numbers from the main body of the paper.

\subsection{Object detection} \label{sec:31}
For ablation purposes, we evaluate our model on COCO~\cite{COCO} to verify our performance on object detection. In Table~\ref{tbl:cocoabl} (a), we evaluate on full-size COCO images, whilst in Tables~\ref{tbl:cocoabl} (b,c) we evaluate on COCO images \textit{transformed} in order to train the tracker.

\noindent\textbf{Location queries: } Although the concept behind semantic queries is intuitive, we demonstrate their effectiveness on the validation split of COCO dataset.
We first train the model for object detection. At training time we feed in perturbed ground truth bounding boxes as queries to the decoder, and during evaluation we use either encoder proposals generated by~\cite{zhu2021ddetr} in their two-stage model, slightly perturbed bounding boxes, encoder-proposals as queries. For upper-bound comparison, we evaluate the model by feeding in the direct ground truth embeddings as queries as well. Table~\ref{tbl:cocoabl}~(a) shows that the model can indeed use the position embedded in the queries to further refine their location. By training the model to refine object proposals, we can even improve upon the two-stage proposal model proposed by~\cite{zhu2021ddetr} when using their encoder proposals. We see that detection becomes an easy task, therefore location-based queries are forced to learn how to retrieve important identifying information about the objects given their location.

\noindent\textbf{Object detection whilst tracking: } During tracking, we are still performing object detection: we might want to initialize new tracks at the initial stage (\textit{init}), or ``pick up" new ones whilst tracking (\textit{tracking}). Therefore, we show the object detection performance of our semantic queries in a simulated tracking scenario. We simulate tracking data for training and evaluation from COCO as described in Sec.~3.2 of the main paper. During the evaluation, we evaluate our initial \texttt{pos} queries for object detection during the init stage (original image), and we show how well various semantic queries perform during the tracking stage (spatially augmented image, where ground truth is augmented as well). During the evaluation, each type of query is fed into the decoder separately (avoiding information leakage) and subsequently matched to the ground truth using the matching process described in~\cite{zhu2021ddetr}. When using \texttt{det} queries, we follow the protocol in~\cite{zhu2021ddetr} by using $N_{\text{object}}=100$ queries. Results in Table~\ref{tbl:cocoabl}~(b) show that training for joint tracking and detection impacts the model's performance on the detection-only task (which is expected, as the regression-head now learns to \textit{predict} the location at the next frame). They also show that the model is able to successfully disentangle the semantic information from various queries -- appearance query knows far less about the object location compared to the ground-truth location one.

\noindent\textbf{Feature aggregation -- object detection: } In the ablation study above, we look at each query independently, but in our final model, we are aggregating embedding information together. Therefore, we investigate four different aggregation mechanisms, still whilst in the setting of object detection. Now in the tracking stage, instead of computing the embedding for each query separately, we pass them through the encoder together, thus allowing the model to attend to both the appearance and the location of the object. Since for each known object, we now produce three common embeddings $V_{t}$ for $t\in\{\texttt{pos}, \texttt{id}, \texttt{both}\}$. We aggregate these embeddings in four different ways: 1) max-pooling, 2) average-pooling, 3) concatenate and pass through an MLP, or 4) pass through collaborative-gating unit~\cite{liu2019use}. Collaborative gating marginally outperforms other aggregation methods. Full results can be found in Table~\ref{tbl:cocoabl}~(c).

\begin{table*}[!hb]
\centering
\caption{Ablation study of semantic queries used for object detection, evaluated on COCO validation set.\\}
\label{tbl:cocoabl}

\scriptsize
\begin{subtable}[t]{.45\linewidth}
\centering
\begin{tabular}{@{}cl|l@{}}
\toprule
Models         & Query                 & AP   \\ \midrule
Baseline dDETR & detection              & 43.8 \\
Baseline dDETR & two-stage proposals & 46.2 \\
Ours           & two-stage proposals          & 49.8 \\
Ours           & perturbed GT bbox            & 71.6 \\
Ours           & GT bbox                      & 75.1 \\ \bottomrule
\end{tabular}
\caption{\scriptsize Comparison of various semantic queries (including ground truth (GT) for the upper bound) when models are trained purely for object detection on COCO. Model learns to effectively use the provided ground truth data, and refine perturbed boxes demonstrating that refinement can be done accurately given precise enough bounding boxes.}
\end{subtable}%
\hspace{4pt}
\begin{subtable}[t]{.24\linewidth}
\centering
\begin{tabular}{@{}cl|l@{}}
\toprule
Query      & Stage & AP   \\ \midrule
GT bbox    & init  & 71.9 \\
\texttt{pos}  & tracking  & 69.1 \\
\texttt{id} & tracking  & 37.4 \\
\texttt{both}       & tracking  & 62.0 \\
\texttt{det}     & tracking  & 42.8
\end{tabular}
\caption{\scriptsize Comparison of various semantic queries trained on tracking data simulated from COCO, and evaluated on object detection. We show that model is capable of object detection even when trained for tracking, and that our method is successful in disentangling the semantic information into separate queries for the tracking step.}
\end{subtable}%
\centering
\hspace{4pt}
\begin{subtable}[t]{.24\linewidth}
\centering
\begin{tabular}{c|l}
\toprule
Aggregation method   & AP   \\ \hline
max-pool            & 54.7 \\
avg-pool            & 61.1 \\
cat + MLP           & 71.1 \\
collaborative gating & 71.5
\end{tabular}
\caption{\scriptsize Comparison of different embedding-aggregation strategies for known objects at tracking stage. We show that learnable aggregation strategies have significant advantage over simple pooling strategies.}
\end{subtable}
\end{table*}

\subsection{Single- vs multi-query tracking} \label{sec:32}
In Table~\ref{abl:querynum}, we further expand upon the the ablation study from the main body of the paper (Table~5~(a)), namely by looking at various permutations of queries. The output of the queries is then aggregated using collaborative-gating aggregation function~\cite{liu2019use} as described in Section~3.1.5 of the paper. 

\begin{table}[!htb]
\centering
\caption{Further analysis of different query-combinations. \\}
\begin{tabular}{llllcc}
\toprule
\texttt{det} & \texttt{id} & \texttt{pos} & \texttt{both} & \multicolumn{1}{c}{MOTA~$\uparrow$} & \multicolumn{1}{c}{IDF1~$\uparrow$} \\ \midrule
\checkmark & \checkmark & \checkmark & \checkmark & 68.3 & 66.1 \\
\checkmark &  &  &  & 55.9 & 54.7 \\
\checkmark & \checkmark &  &  & 57.4 & 58.2 \\
\checkmark &  & \checkmark &  & 61.1 & 57.4 \\
\checkmark &  &  & \checkmark & 62.9 & 63.5 \\
 & \checkmark &  &  & 52.7 & 54.1 \\
 & \checkmark & \checkmark &  & 59.4 & 58.6 \\
 & \checkmark &  & \checkmark & 58.2 & 57.0 \\
 & \checkmark & \checkmark & \checkmark & 65.1 & 64.8 \\
 &  & \checkmark &  & 56.1 & 55.8 \\
 &  & \checkmark & \checkmark & 61.6 & 64.1 \\
 &  &  & \checkmark & 56.9 & 56.2 \\ \bottomrule
\end{tabular}
\label{abl:querynum}
\end{table}

We can see two broad trends. First,  that including \texttt{det} queries during tracking significantly increases the performance of our model. Second, that \texttt{pos} and \texttt{both} queries tend to carry more relevant information for tracking purposes compared to \texttt{id} queries. 


\subsection{Heuristic method for query aggregation} \label{sec:33}
In the main paper (Table~6~(b)), we show the efficacy of various output aggregation schemes and compare them to, what we refer to as, the heuristic method.
Since most tracking-by-detection systems have two distinct tasks (detection and data-association), we define the heuristic method as using the output corresponding to a particular query type for each task. In Table~\ref{abl:perm}, we present all permutations of queries that in the end lead to the result presented in Table~6~(b) of the main paper, here shown in bold.
Note that, unlike in Table~\ref{abl:querynum}, \texttt{det} queries are present in all permutations and we void the query aggregation scheme completely.

\begin{table}[]
\centering
\caption{Further analyses of the heuristic method in Table 6b of the main paper. Instead of aggregating the outputs of semantic queries, we use a specific one for each stage of tracking process. Note that \texttt{det} queries are present in all cases. The last row shows performance of detection-queries only (i.e.\ model without explicit auto-regressive information propagation).\\}
\begin{tabular}{@{}llcc@{}}
\toprule
Detection & Data-association & MOTA~$\uparrow$ & IDF1~$\uparrow$ \\ \midrule
\texttt{pos} & \texttt{pos} & 59.4 & 54.9 \\
\texttt{pos} & \texttt{id} & \textbf{65.7} & \textbf{64.9} \\
\texttt{pos} & \texttt{both} & 63.8 & 61.3 \\
\texttt{id} & \texttt{pos} & 56.9 & 52.2 \\
\texttt{id} & \texttt{id} & 55.7 & 55.1 \\
\texttt{id} & \texttt{both} & 55.4 & 53.8 \\
\texttt{both} & \texttt{pos} & 58.8 & 53.4 \\
\texttt{both} & \texttt{id} & 64.6 & 64.3 \\
\texttt{both} & \texttt{both} & 64.8 & 64.6 \\ \midrule
\texttt{det} & \texttt{det} & 53.1 & 50.4\\
\bottomrule
\end{tabular}
\label{abl:perm}
\end{table}

Firstly, we can see that auto-regressive information propagation is key for good performance as the performance of \texttt{det} queries is lower than, for example, auto-regressively propagated \texttt{det} queries by $8.3$ MOTA points. Furthermore, our findings reinforce those by~\citet{yu2021relationtrack}. While they use a learned module to disentangle detection and re-identification information, our embeddings are disentangled by separate feed-forward networks (appearance head, bounding-box detection head), we find that using specialised embeddings for each step outperforms using joint embeddings -- in our case by $0.9$ MOTA points with no aggregation, and $3.8$ MOTA points when a learned aggregation module is applied.

\section{Further performance enhancements via offline tracking} \label{sec:4}

In Table~1 of the main paper, we RelationTrack~\cite{yu2021relationtrack} stands out compared to all other methods. Indeed, they report their baseline model, which shares the same backbone as MQT, achieves 68 MOTA points~\footnote{There is no specification on the evaluation protocol used; we assume standard 7-way split cross-validation.} (compared to 56 for MQT). A potential reason for such improvement is their tracking protocol, which consists of offline refinement of the initially proposed tracklets and further post-processing including trajectory-filling strategies discussed in~\cite{han2020mat}.

As our model doesn't have a separate data-association stage we instead attempt to imitate this two-step procedure by doing two separate tracking passes on the video: one from beginning to the end, and another from the end to beginning. If the track bounding boxes overlap by over 50\% of the total area, we consider the tracks to be true positives. We refer to this as back-to-front track confirmation (B2F track). Additionally, we attempt to use the Hungarian algorithm with a matching threshold of 0.4 to match tracks from a beginning-to-end pass with the appearance vectors of an end-to-beginning pass and vice-versa and repeat the overlap process from above. We refer to this as back-to-front appearance confirmation (B2F id). Lastly, we adopt the trajectory-filling (tf) strategy as in~\cite{han2020mat, yu2021relationtrack} on final predicted tracks. 

\begin{table*}
\caption{Comparing the effects of various offline post-processing enhancements on our model on a standard 7-way split cross-validation for MOT17. Our model uses detections from~\cite{yolox2021} in ``Private+" setting. \\}
\label{tbl:tracklets}
\scriptsize
\begin{subtable}[t]{.45\linewidth}
\centering
\begin{tabular}{@{}lccccc@{}}
\toprule
Method & B2F track & B2F id & TF & MOTA & IDF1 \\ \midrule
\multicolumn{4}{l}{RelationTrack--baseline~\cite{yu2021relationtrack}} & 68.5 & 73.3 \\ 
Ours &  &  &  & 56.9 & 56.2 \\
Ours & \checkmark &  &  & 60.1  & 59.8 \\
Ours & \checkmark & \checkmark &  & 62.7 & 62.0 \\
Ours & \checkmark & \checkmark & \checkmark & 63.4 & 64.2 \\ 
Ours &  &  & \checkmark & 58.0 & 58.6 \\\bottomrule
\end{tabular}
\caption{\scriptsize Effects of the performance-enhancing post-processing methods on our baseline model using only \texttt{both} track queries.}
\end{subtable}%
\hspace{4pt}
\begin{subtable}[t]{.45\linewidth}
\centering
\begin{tabular}{@{}lccccc@{}}
\toprule
Method & B2F track & B2F id & TC & MOTA & IDF1 \\ \midrule
\multicolumn{4}{l}{RelationTrack~\cite{yu2021relationtrack}} & 70.2 & \textbf{75.3} \\
Ours &  &  &  & 68.3 & 66.1 \\
Ours & \checkmark &  &  & 69.7 & 69.8 \\
Ours & \checkmark & \checkmark &  & 71.6 & 71.1 \\
Ours & \checkmark & \checkmark & \checkmark & \textbf{72.1} & 73.3 \\ 
Ours &  &  & \checkmark & 68.9 & 72.1 \\\bottomrule
\end{tabular}
\caption{\scriptsize Effects of the performance-enhancing post-processing methods on our best model. Note that semantic queries and our multi-query tracking protocol significantly reduce the need for an additional offline post-processing.}

\end{subtable}
\end{table*}

\section{Comparison with similar methods} \label{sec:5}

Several related works concurrently attempt to adapt the transformer architecture to the MOT problem. Most of these works differ from \textit{MQT} either in the tracking paradigm, or in architectural details. Bellow, we outline differences to the most similar work. The reader should note however, that \textit{none} of these concurrent works can be generalised to the class agnostic GMOT or TAO tasks as they use tracking-by-detection approaches.

\textbf{TransCenter}~\cite{xu2021transcenter} shares architectural similarities with \textit{MQT} (in that they use a deformable transformer and semantically separable queries), but the finer details and their tracking paradigm are fundamentally different. On an architectural level, they utilise query leaning networks to separate queries from the encoder representation. Furthermore, they utilise \textit{two separate} decoder modules, one for each query type. We feed all the queries concurrently in the single decoder. More importantly, their tracking paradigm is fundamentally different as they compute \textit{dense} queries for location and tracking displacement (as opposed to appearance like us). They show that given dense detection and tracking memory, tracking can emerge from these inputs alone. To this end, they design and combine custom decoder modules in order to aid in matching the model outputs over time. Their paradigm renders the reID module unnecessary, but it comes at the expense of higher model complexity. While clearly innovative and effective,  their approach performs better than \textit{MQT} on one metric (MOTA 71.9 vs 65.4) but worse on others (IDF1 62.3 vs 63.4, or ID switches 4626 vs 1104). \\
\textbf{MOTR}~\cite{zeng2021motr} shares the transformer backbone with \textit{MQT}, and they also leverage the power of semantically decoupled embeddings. Unlike us, however, their embeddings are decoupled via ``global context disentangling unit" from the final layer output of a backbone CNN. We find that we can do it implicitly in the transformer decoder.\\
\textbf{Trackformer}~\cite{trackformer} is most similar in spirit to our \textit{MQT} method. Their detection queries are analogous to our static \texttt{det} queries, and their tracking queries are analogous to our \texttt{both} queries. In the ablation studies (Tbl~5~(a); Tbl~3 in supp.\ material) we show the effectiveness of using multiple (and semantically decoupled) queries over the single tracking query paradigm of~\cite{trackformer}.\\

\subsection{Running speed comparison} \label{sec:51}
At the time of writing, only~\cite{trackformer} code was available for a direct comparison. In our experiments, we find that pre-training \textit{MQT} on the CrowdHuman dataset takes 8 days using 6 V100 GPUs, which is comparable to the original implementation of~\cite{trackformer} on similar hardware.\\

\bibliographystyle{abbrvnat}
{\footnotesize
\bibliography{egbib}}

\end{document}